\documentclass[letterpaper, 10 pt, conference]{ieeeconf}  

\IEEEoverridecommandlockouts                              

\usepackage{graphics,caption} 
\graphicspath{ {figure/} }
\usepackage{graphicx} 
\usepackage{epsfig} 
\usepackage{times} 
\usepackage{amsmath} 
\usepackage{amssymb} 
\usepackage{cite}
\usepackage{booktabs}
\usepackage{threeparttable}
\usepackage{multirow}
\if CLASSOPTIONcompsoc
\usepackage[caption=false, font=normalsize, labelfont=sf, textfont=sf]{subfig}
\else
\usepackage[caption=false, font=footnotesize]{subfig}
\usepackage{url}
\usepackage{mathrsfs}
\usepackage[colorlinks,linkcolor=blue]{hyperref}

\usepackage{float}
\usepackage{stfloats}

\title{\LARGE \bf
Frontier Semantic Exploration for Visual Target Navigation\\
}

\author{Bangguo Yu, Hamidreza Kasaei, Ming Cao
\thanks{This work of Yu is supported in part by the China Scholarship Council.}
\thanks{All authors are with the Faculty of Science and Engineering, University of Groningen, 9747 AG Groningen, the Netherlands. {\tt\small \{b.yu, hamidreza.kasaei, m.cao\}@rug.nl}}%
}

\begin{document}

\maketitle
\thispagestyle{empty}
\pagestyle{empty}

\begin{abstract}

	This work focuses on the problem of visual target navigation, which is very important for autonomous robots as it is closely related to high-level tasks. To find a special object in unknown environments, classical and learning-based approaches are fundamental components of navigation that have been investigated thoroughly in the past.
	However, due to the difficulty in the representation of complicated scenes and the learning of the navigation policy, previous methods are still not adequate, especially for large unknown scenes.
	Hence, we propose a novel framework for visual target navigation using the frontier semantic policy.
	In this proposed framework, the semantic map and the frontier map are built from the current observation of the environment. Using the features of the maps and object category, deep reinforcement learning enables to learn a frontier semantic policy which can be used to select a frontier cell as a long-term goal to explore the environment efficiently.
	Experiments on Gibson and Habitat-Matterport 3D (HM3D) demonstrate that the proposed framework significantly outperforms existing map-based methods in terms of success rate and efficiency. Ablation analysis also indicates that the proposed approach learns a more efficient exploration policy based on the frontiers. A demonstration is provided to verify the applicability of applying our model to real-world transfer.
	The supplementary video and code can be accessed via the following link: \href{https://sites.google.com/view/fsevn}{https://sites.google.com/view/fsevn}.

\end{abstract}

\section{INTRODUCTION}

Imagine a robot is walking into your room, and you say: ``Hi, robot, can you help me check if my laptop is in my bedroom? If so, bring it to me.'' In such a high-level task, the first idea of the robot is to find the laptop in your bedroom. If the model of the scene is known, the robot can just plan a path to the location of the laptop. However, if the location has changed or is unknown, this task would be a challenge for the robot.
Visual target navigation is one of the important functionalities for intelligent robots to execute more high-level tasks. To find an object as the appointed category in an environment, the robot has to be equipped with sufficient abilities of scene understanding and autonomous navigation.
The indoor environment is usually complicated and labile, which requests the robot not only to find the target but also to maintain the great generalization in unknown scenes.

The task of visual target navigation has attracted much research attention.
Most of the classical approaches focus on building the model of the scene and planning the explicit path based on the model, which is limited by unexplored or changing scenes. To achieve better performance in this task, more precise semantics and geometry information are also required.
With the development of machine learning, learning-based methods are used increasingly for this task. For the end-to-end models, the first framework\cite{Zhu2017} uses deep reinforcement learning to find an object as the appointed category in an unknown environment based on current observable images. Later, the framework is expanded by many researchers to improve the navigation performance\cite{Yang2019}\cite{Lyu2022}\cite{Druon2020}\cite{Ye2021a}. On the other hand, map-based models are also applied in this task and achieve satisfying performances compared with end-to-end methods. The map-based method is demonstrated in \cite{Chaplot2020b}, which attempts to build partial map based on current observation as the long-term memory and learns the semantic priors using deep reinforcement learning to select the waypoint. The key to those learning-based approaches is the simulation environment. To date, many excellent simulation platforms have been proposed, such as AI2-THOR \cite{Zhu2017}, AI Habitat \cite{Savva2019}, and GibsonEnv \cite{Xia2018}.

\begin{figure*}[htbp]
	\centering
	\includegraphics[scale=0.53]{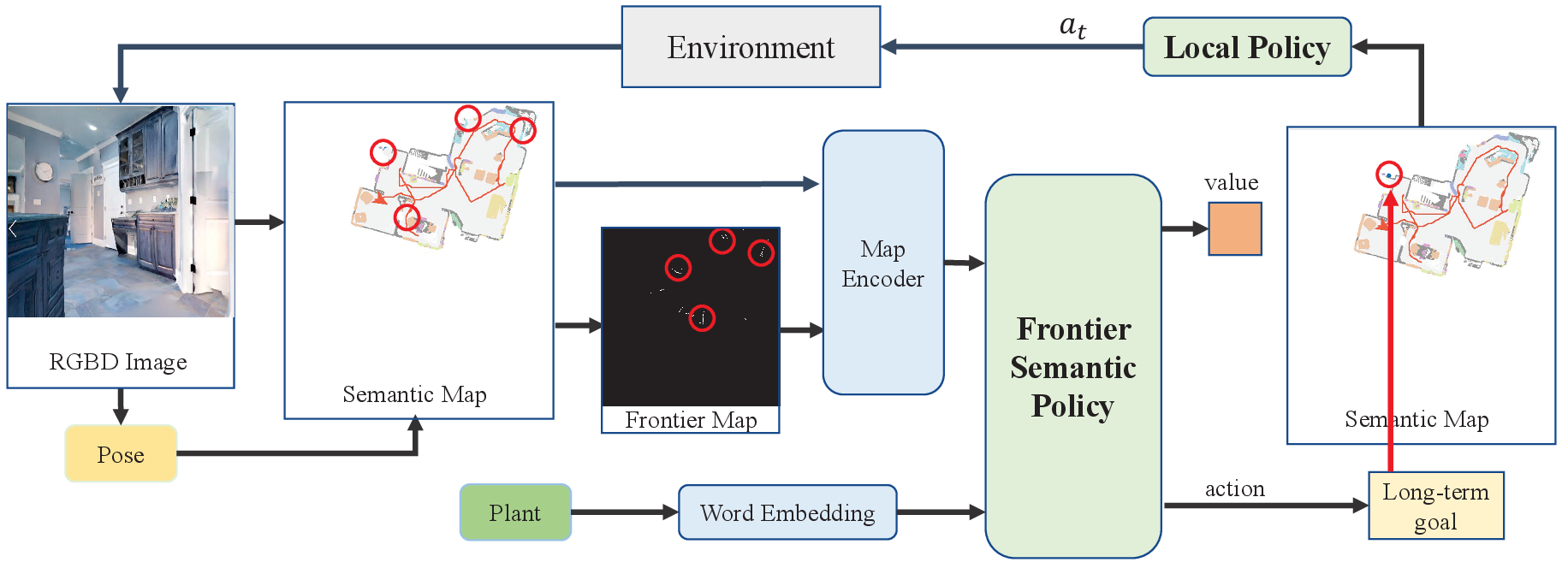}
	\caption{The overall architecture of the target navigation framework. This framework uses as input the RGB-D images to generate the semantic and frontier maps, and selects the long-term goal based on the maps and object category via deep reinforcement learning. After reaching the long-term goal, the local policy then guides the final action for the robot.}
	\label{fig:system_architecture}
\end{figure*}

In this work, we focus on the problem of navigating to find an object as the appointed category in an environment based on the visual observation. We propose to achieve an efficient exploration and searching policy using the frontiers and semantic map features. The overall framework is shown in Fig \ref{fig:system_architecture}. Visual observation is used to construct the real-time semantic map and extract the frontiers. Based on the frontier and semantic maps, deep reinforcement learning is adopted to learn the navigation policy. We also use the invalid action masking method \cite{Huang2022} to deal with the problem that the number of the frontiers is changing during the navigation. We evaluate our model on the simulation platform Habitat \cite{Savva2019} and compare the navigation performance against previous map-based approaches. We perform experiments on the photorealistic 3D environments of Gibson \cite{Xia2018} and HM3D\cite{Ramakrishnan2021a}. In sharp contrast to many other map-based methods, our framework integrates the features of the semantic and frontier maps into the policy, and selects the long-term goal from the frontiers. Ablation experiments show the effectiveness of the frontier-based policy. We also demonstrate the application of our method in real-world settings.

Our contributions are summarized as follows:

\begin{itemize}

	\item We propose a framework that can build the environment map and select the long-term goal based on the frontiers using deep reinforcement learning to achieve efficient exploration and searching.
	\item We show that using the invalid action masking method to learn to select the long-term goal from frontiers, the policy can significantly improve the performance.
	\item We apply our model in the robot platform to verify the performance in the real world, and discuss the gap of sim2real for the visual target navigation task.

\end{itemize}

\section{RELATED WORK}

Visual target navigation is one of the most fundamental tasks for intelligent robots, which is inspired by human behaviors for object searching. We discuss related works of classical and learning-based methods.

\subsection{Classical Semantic Navigation}

Classical approaches for visual target searching focus on mapping and navigation. After planning the path based on the geometric and semantic map, the target navigation can be seen as an obstacle avoidance problem. \cite{Labbe2013} presented an approach of building the map for 3D environment and then the path can be planned based on the map. \cite{Nakajima2019}\cite{Sun2018}\cite{Grinvald2019} incorporated the semantic information into the object-centric map, which can help to understand the scene.
The accuracy of the semantic map has a great influence on the performance of the navigation. The main limitation faced by classical methods is the unexplored or changing scenes since the map cannot be built in advance.

\subsection{Learning-based Semantic Navigation}
\subsubsection{End-to-end model}
After \cite{Zhu2017} proposed the end-to-end framework for visual target navigation, there has been a surge of novel approaches to this problem using end-to-end deep reinforcement learning methods.
\cite{Zhu2017} used pre-trained ResNet \cite{He2016} to encode the input observation and target image, and fused them into the Asynchronous Advantage Actor-Critic (A3C) \cite{Mnih2013} model. To train the navigation policy, the simulation platform AI2-THOR was proposed by \cite{Zhu2017}, which provides a photorealistic environment with high-quality 3D scenes. This work confirmed the key of the learning-based method for the target navigation task. Hence, \cite{Yang2019} proposed using Graph Convolutional Network (GCN) \cite{Kipf2017} to incorporate the prior knowledge into a deep reinforcement learning framework. \cite{Lyu2022} used attention in the 3D knowledge graph after encoding the images by GCN. \cite{Druon2020} proposed the self-defined context grid rather than using the existing architecture of the network to encode the visual feature and find the relation between the objects.
\cite{Du2020} combined the object relation graph and other methods to improve the performance of target navigation. \cite{Ye2021a} learned auxiliary tasks such as predicting agent dynamics, environment states, and map coverage simultaneously to improve the performance. \cite{Maksymets2021} used data augmentation to address overfitting problem.


\subsubsection{Map-based model}
Many end-to-end methods used past several images \cite{Zhu2017}\cite{Yang2019}\cite{Druon2020} or RNN \cite{Lyu2022}\cite{Druon2020}\cite{Wijmans2019}\cite{Mousavian} to remember the scene features.
\cite{Druon2020} had reported that in the navigation tasks the RNN may work better than the method of stacking past features, but in \cite{Fang2019} the use of RNN is not recommended because RNN often fails to capture long-term dependencies.
The map-based method has a great advantage in that the map can be used as the long-term memory.
So \cite{Chaplot2020} proposed a new framework to build the map by networks and learn the policy of the global goal to guide the agent to explore the environment, which achieves great efficiency in point-to-point navigation. \cite{Chaplot2020a} constructed the topological graph to represent the scene and calculate the matching score for each node based on the goal image and source image. \cite{Chang2020} built the same topological graph node representing images from YouTube videos and used Q-learning to score each node. \cite{Ramakrishnan2022} used the supervised learning to predict the frontier as the long-term goal, which also achieved great performance.

Closely related to the approach presented in this paper is the recent work in \cite{Niroui2019} and \cite{Chaplot2020b}, with the similar aim of exploring the unknown environment based on the partial map feature. \cite{Niroui2019} combined the traditional approach of frontier-based exploration\cite{Yamauchi1997} with deep reinforcement learning to allow the robot to explore the unknown environment. \cite{Chaplot2020b} incorporated the explicit semantic map with \cite{Chaplot2020} to learn the semantic context policy for efficient target navigation. Our method addresses the problem of learning a frontier-based navigation policy based on frontiers and semantic map features for visual target navigation.

\section{The Proposed Method}

In this section, we describe the definition of the target navigation task and the main modules of our framework.

\subsection{Task Definition}

In the visual target navigation task, the agent should navigate the environment to find an object as the appointed category in a scene. The category can be mapped from a predefined set $C = \left\{c_0, \dots, c_m\right\}$ and the scene can be represented by $S = \left\{s_0, \dots, s_m\right\}$.
For each task, the agent is initialized at a random position $p_i$ in the scene $s_i$ and receives the target object category $c_i$, so the task can be denoted by $T_i = \left\{s_i, c_i, p_i\right\}$. No ground-truth map is available from the scene.
At each time step $t$, the agent would receive an observation $o_t$, take action $a_t$ and get a scalar reward $r_t$, so we can define it as the Partially Observable Markov Decision Process\cite{Nanni2011}. The observation contains RGB-D images, the location and orientation of the agent, and the object category. The action space $\mathcal{A}$ consists of four actions: $move\_forward, turn\_left, turn\_right$, and $stop$. Each $move\_forward$ makes the agent move 25 $cm$. $turn\_left$ and $turn\_right$ can rotate the agent by 30 degrees. The $stop$ action means that the agent knows when it is close to the target. If the distance between the agent and target is less than 0.1m and the agent takes the $stop$ action, the episode task is considered successful. The maximum number of timesteps in one episode is 500.

\subsection{Overview}

As illustrated in Fig \ref{fig:system_architecture}, our framework uses deep reinforcement learning to select the long-term goal based on the semantic and frontier maps.
Firstly, the agent obtains the observation of the environment to build the semantic map. The frontier map is extracted from the explored map and obstacle map, and the target is encoded by word embedding.
Secondly, the frontier semantic policy would decide a long-term goal from all the frontiers based on the current observation. After getting the long-term goal, the local policy would plan a path and take the action to explore the environment and search for the target.

\subsection{Map Representation}

\subsubsection{Semantic Map}

When the sensory input, the semantic map is built based on RGB-D images and the position of the agent. We use a similar representation as \cite{Chaplot2020b} to denote the semantic map as a $ K \times M \times M $ matrix, where $M \times M$ denotes the map size, and $K = C_n+2$ is the number of channels in the semantic map. $C_n$ is the number of semantic categories, and the first two channels represent the obstacle and explored map. The map is initialized with all zeros at the beginning of each episode, and the agent starts at the center of the map.

To build the environment map, we use the geometric method that transforms the visual input into the 3D cloud points, and then projects the point cloud into a top-down 2D map based on the agent's location. Specifically, the obstacle and explored map channels depend on the depth image, and other channels of the map are projected by the output of semantic segmentation. We select a height range of the point cloud to generate the obstacle map and take all of the point projections as the explored map. Based on the output of semantic segmentation, the corresponding cloud points are aligned and each channel of the semantic mask is projected into the proper position of the semantic map.

A big challenge of the map-based method for the visual target navigation task is the multi-floor scenes. In some cases, the object goal and the agent's initial position are not on the same floor. To find the object, the agent must walk upstairs or downstairs on different floors, which would make a lot of influence on the 2D map building. To deal with this problem, we remove the stairs area in the obstacle map so the agent can cross them. We also update the map to clear the old floor obstacle when the agent comes into a new floor.

\subsubsection{Frontier Map}

\begin{figure}[htbp]
	\centering
	\includegraphics[scale=0.35]{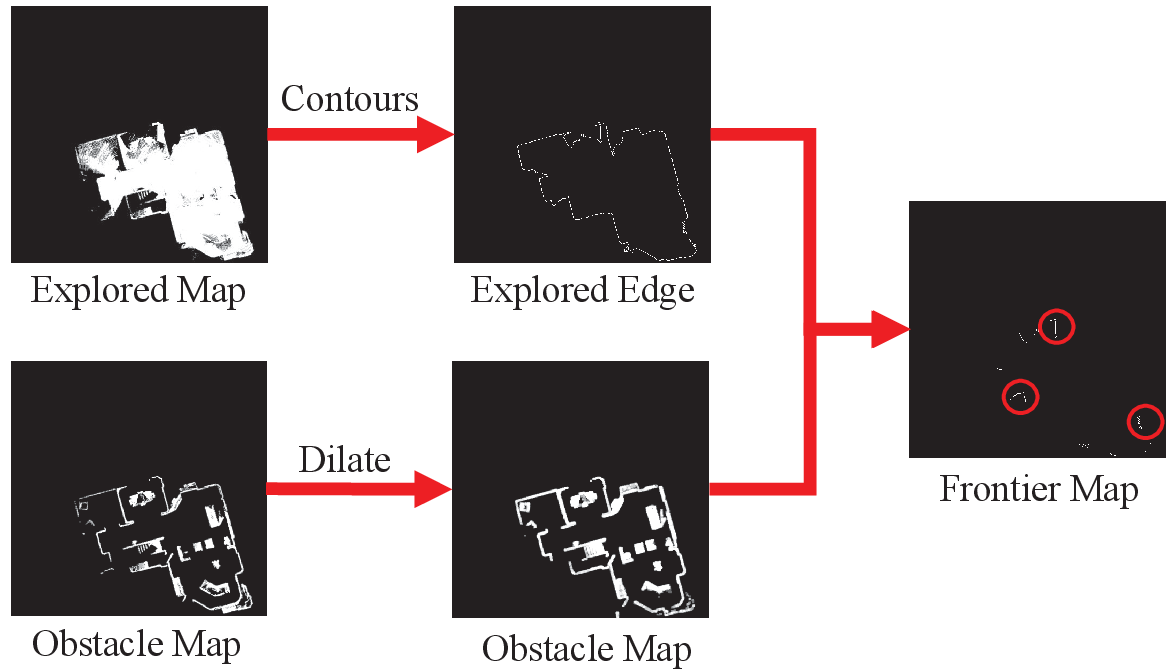}
	\caption{The process of generating the frontier map based on the explored map and obstacle map.}
	\label{fig:frontier-map}
\end{figure}

The frontier map is generated from the explored map and obstacle map. As illustrated in Fig \ref{fig:frontier-map}, the explored edge is extracted by finding maximum contours from the explored map, and after dilating the edge of the obstacle map, the frontier map can be represented as the differences between explored and obstacle maps.
Frontier cells are identified and clustered in chains by connected neighborhood. Clusters that are too small are removed. The cost-utility \cite{Julia2012} approach is used to select the top 4 score frontier cells from the frontier map. We use $ 4 \times M \times M $ to represent the frontier map, and each channel just contains one cell.
The cells in the center of the remaining cluster chains compose the subset of candidate destinations $F$. So for each frontier cell $a \in F$, we can get the score $S^{C U}(a)$
\begin{equation}
	S^{C U}(a)=U(a)-\lambda_{C U} C(a)
\end{equation}
where $U(a)$ is a utility function, $C(a)$ is a cost function and the constant $\lambda_{C U}$ adjusts the relative importance between both factors.
Then, the target frontier cell $t$ is chosen as the one that maximizes the utility-cost function:

\begin{equation}
	t=\underset{a \in F}{\arg \max } S^{C U}(a) .
\end{equation}

\subsection{Policy Learning}

Due to the great advantage of the map-based method, global and local policies are applied in visual target navigation. The global policy decides the long-term goal from the frontiers based on the current observation to guide agents to find the target object. The local policy is used to plan a path and output the final action of the agent based on the long-term goal.

\subsubsection{Global Policy}

We use the Proximal Policy Optimization (PPO) algorithm \cite{Schulman2017} to learn the frontier semantic policy based on the maps and object categories. In there we focus on the design of the state space and action space for the learning of PPO in visual target navigation.

\begin{figure}[htbp]
	\centering
	\includegraphics[scale=0.6]{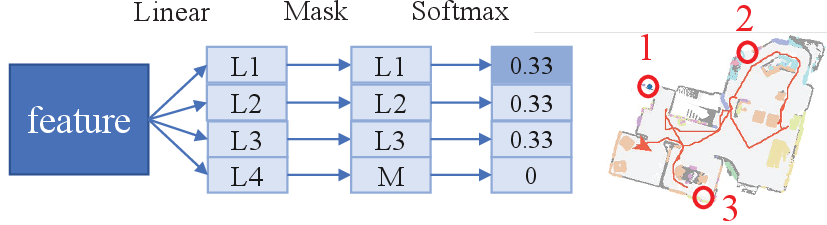}
	\caption{The selection of the frontier cells using invalid action masking. For example, if we get three frontier cells from the observation, the fourth action of action space would be masked to keep the same number as the cells.}
	\label{fig:action-space}
\end{figure}

\begin{itemize}
	\item \textbf{State Space: } The input of the global policy is the joint representation of the current semantic map, frontier map, and object category. We concat all the channels of the semantic map and frontier map to $K+4$ channels. Specifically, to reduce the memory and computing resources, we just crop a slide window on the $K+4$ channels map centering on the location of the robot as the current observation. To extract features from the maps, we use a fully connected layer after 5-layer fully convolutional network to output 2048-d features.

	\item \textbf{Action Space: } The action space of frontier semantic policy is the frontier cells extracted from current observation. Different from \cite{Niroui2019}, the action space is a set of 4 possible actions, which is the same as the number of frontier channels. To avoid sampling the empty frontier channel as long-term goal when the number of frontier cells is less than 4, we use the invalid action masking method\cite{Huang2022} to mask the invalid channel of the frontiers map. The process of invalid action masking can be considered as a state-dependent differentiable function applied for the calculation of policy, which means that the policy gradient update would not be influenced. The process is shown in Fig \ref{fig:action-space}, in which each layer of the action is the index of the frontier map channels. Finally, we use the centroid of the connected domain of sampled frontier cell as the long-term goal.

\end{itemize}

Our implementation of the PPO model contains two fully connected layers, and then the action and value are generated by two branches of the network. We use the frame skipping method under which the global goal is updated every 25 steps, which can improve training efficiency. At each step, if the semantic channel corresponding to the category of the target has non-zero elements, meaning that the target is visible, then all the non-zero elements would be selected as the long-term goal. If the target is invisible, the policy would sample a long-term goal where the target is most likely to be found based on the current semantic and frontier map.

\subsubsection{Local Policy}

After obtaining the long-term goal, the Fast Marching Method (FMM) \cite{Sethian1996} is used to plan a path from the current location of the agent to the goal. A local goal would be selected from the limited range of current positions, and the final action $a_t \in \mathcal{A}$ is taken to reach the local goal. At each step, the local map and local goal are updated using the new observation.
Compared with the end-to-end approach, local policy deals with the problem of learning the ability of obstacle avoidance from scratch, which significantly improves training efficiency.

\subsubsection{Reward Shaping}

On the premise of finding the target, we also focus on minimizing the trajectory length between the start position and the goal. To avoid the sparse reward, three factors are considered: geodesic distance \cite{Anderson2018} to goal, time penalty, and smooth-coverage \cite{Ramakrishnan2021}. At each step $t$, the agent is penalized with a small time term $\lambda$ = -0.001. We use the difference between the current and last geodesic distance as the dense reward:
\begin{equation}
	r_{d} \propto (d_{t-1}-d_{t})
\end{equation}
To encourage the robot to explore the unknown area but reduce the repeating path, we grid the map and incorporate ideas from the smooth-coverage reward \cite{Ramakrishnan2021} to design a shaped version of exploration:
\begin{equation}
	\label{equ:sc}
	\begin{aligned}
		 & r_{e\_t} = \sum_{\mathbf{A}_{i} \in G_{t}} \mathbf{A}_{i} \frac{1}{\sqrt{n_{i}}} \\
		 & r_{\text {e }} \propto r_{e\_t} - r_{e\_t-1}
	\end{aligned}
\end{equation}
where $G_{t}$ represents all the cells in the map at time $t$, $\mathbf{A}_{i}$ is the area explored by the agent in the grid cell $i$, $n_i$ is the number of times the grid $i$ has been observed so far. Each time the robot scans the grid, the number of times of this grid would increment by one. Based on reward equation \ref{equ:sc}, the robot can navigate across less frequently visited grids to explore the unknown area as more as possible. So our final reward function is the sum of all three terms:

\begin{equation}
	r_{t}= r_d+r_{\text {e }}+\lambda
\end{equation}

\renewcommand\arraystretch{1.4}
\begin{table*}[htbp]
	\centering
	\fontsize{10}{9}\selectfont
	\begin{threeparttable}
		\caption{Results of Comparative Study.}
		\label{tab:performance_comparison}
		\setlength{\tabcolsep}{5mm}{}
		{
			\begin{tabular}{ccccccc}
				\toprule
				\multirow{2}{*}{Method}                   & \multicolumn{3}{c}{ Gibson } & \multicolumn{3}{c}{ HM3D }  \cr
				\cmidrule(lr){2-4} \cmidrule(lr){5-7}
				                                          & Success                      & SPL                             & DTG         & Success     & SPL         & DTG\cr
				\midrule
				Random Walking                            & 0.030                        & 0.030                           & 2.580       & 0.000       & 0.000       & 7.600          \cr
				Frontier Based Method \cite{Yamauchi1997} & 0.417                        & 0.214                           & 2.634       & 0.237       & 0.123       & 5.414             \cr
				Random Sample on Map                      & 0.544                        & 0.288                           & 1.918       & 0.300       & 0.143       & 4.761          \cr
				SemExp \cite{Chaplot2020b}                & 0.652                        & 0.336                           & 1.520       & 0.379       & 0.188       & 2.943             \cr
				Frontier Semantic Exp (Our)               & {\bf 0.715}                  & {\bf 0.360}                     & {\bf 1.350} & {\bf 0.538} & {\bf 0.246} & {\bf 3.745}       \cr
				\bottomrule
			\end{tabular}
		}

	\end{threeparttable}
\end{table*}

\begin{figure*}[htbp]
	\centering
	\subfloat[step 1]
	{
		\includegraphics[width=4.2cm]{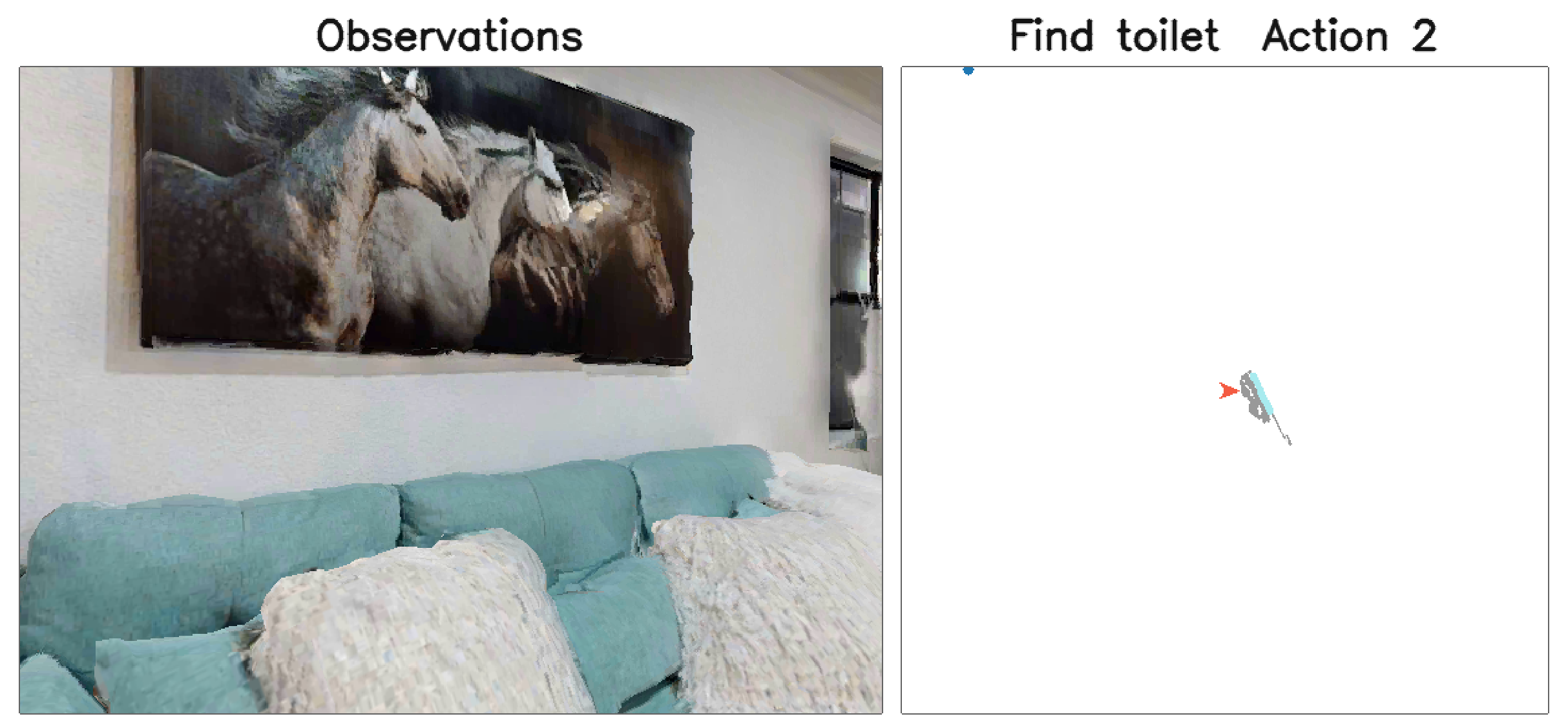}
	}
	\subfloat[step 100]
	{
		\includegraphics[width=4.2cm]{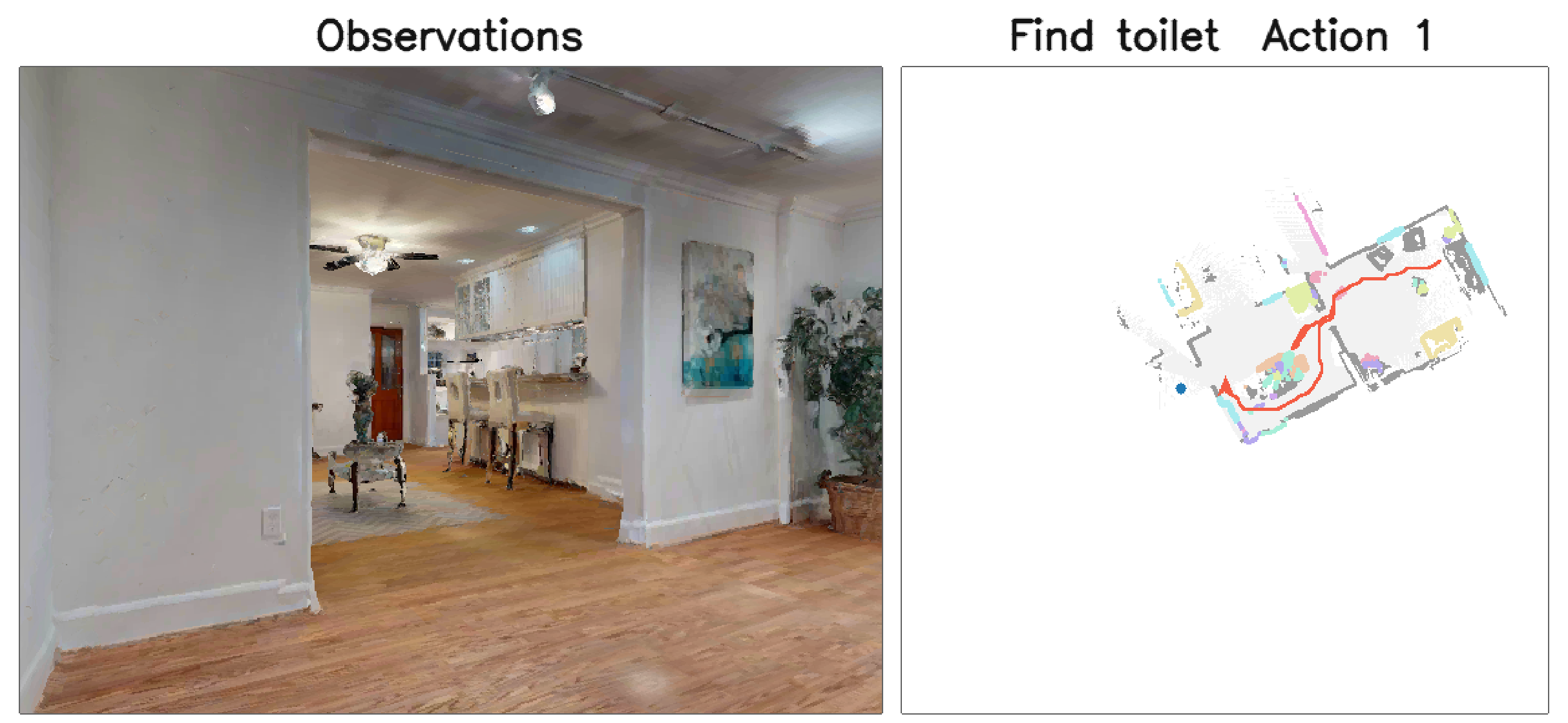}
	}
	\subfloat[step 200]
	{
		\includegraphics[width=4.2cm]{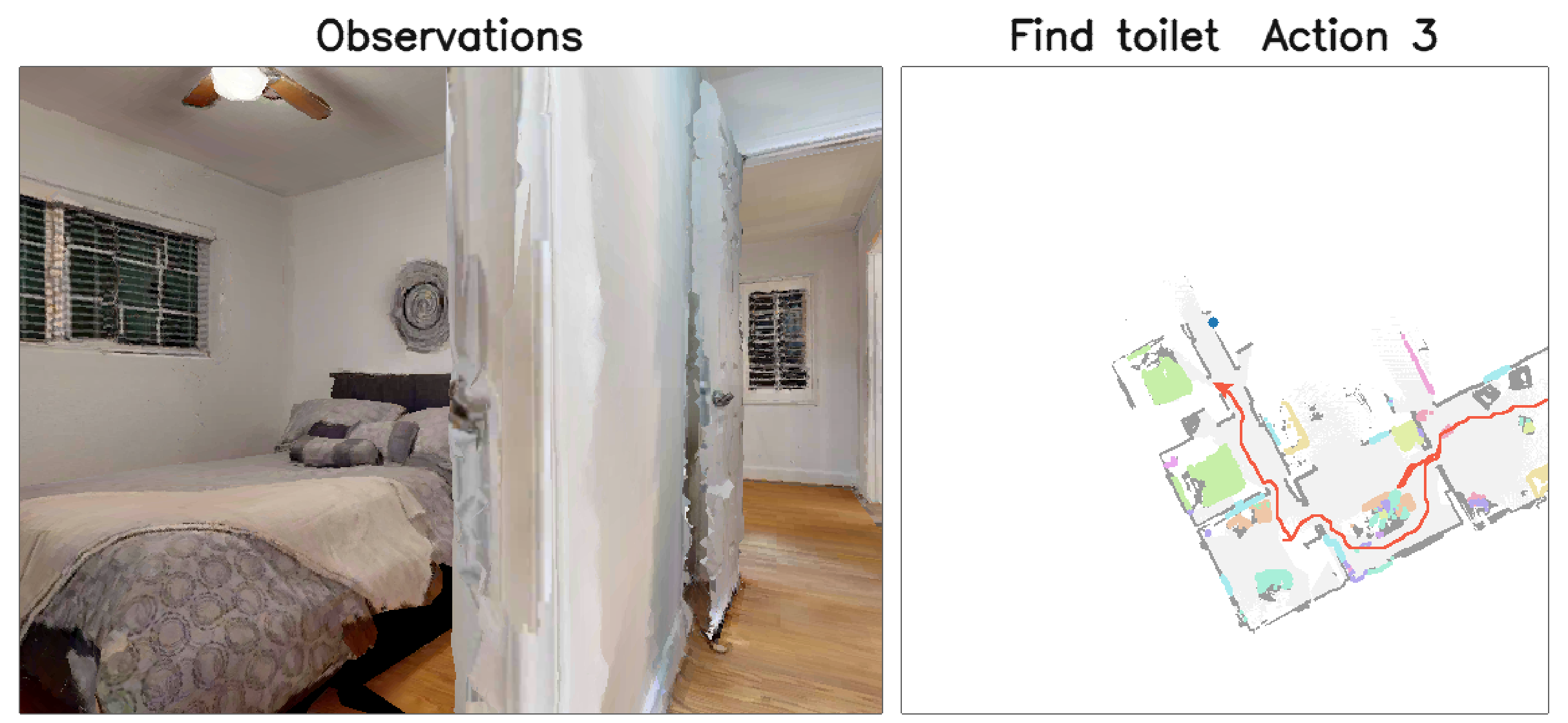}
	}
	\subfloat[step 233]
	{
		\includegraphics[width=4.2cm]{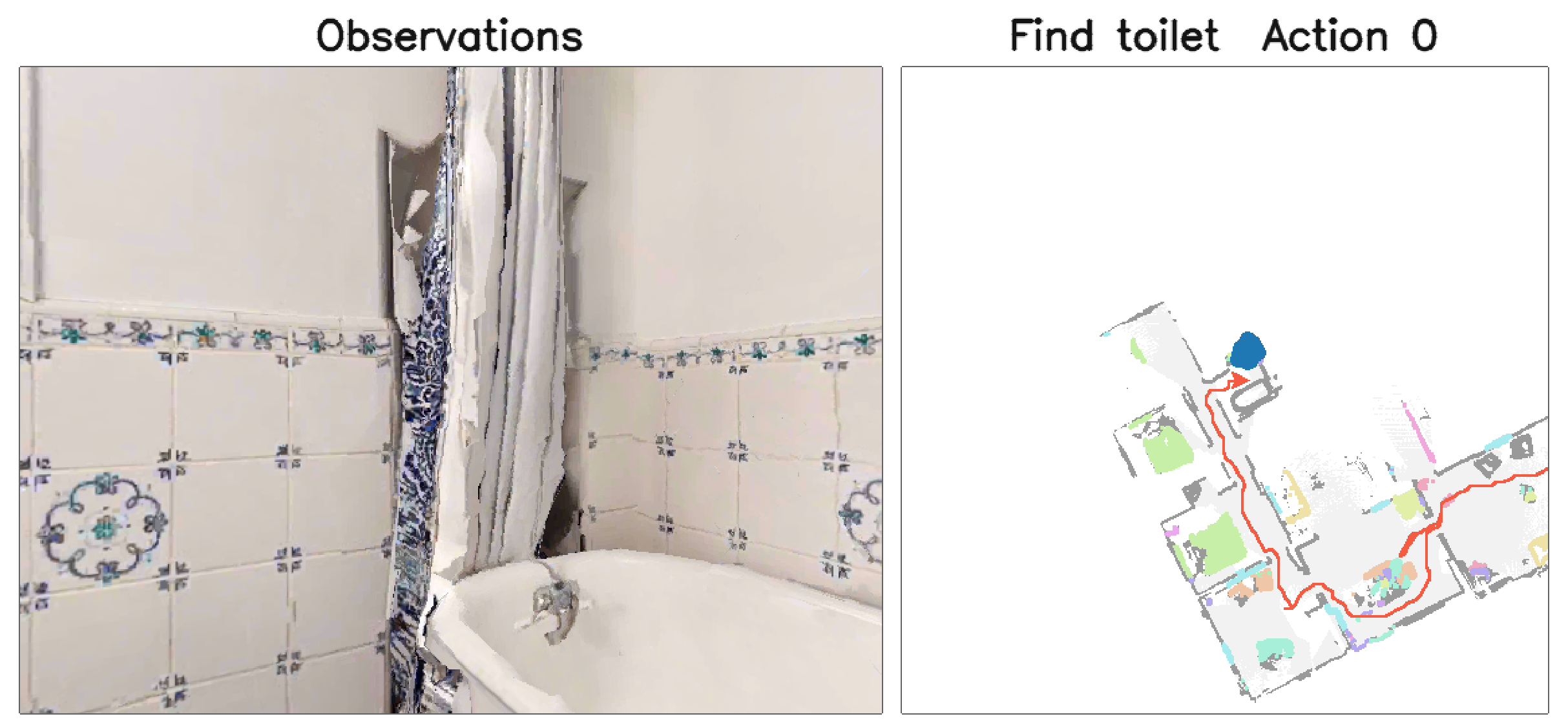}
	}
	\caption{The process of visual target navigation experiment in the Habitat platform to find a toilet. The gray channel represents the barrier, the blue spot shows the long-term goal selected by policy, the red line represents the trajectory of the robot, and other colors represent the semantic objects in the indoor scene.}
	\label{fig:sim_episode}
\end{figure*}

\section{EXPERIMENTS}

In this section, we compare our method with other map-based baselines in the simulation to evaluate the performance of the frontier semantic policy. We also apply our method in the robot platform to verify the application in real-world tasks.

\subsection{Simulation Experiment}
\subsubsection{Dataset}

We perform experiments on HM3D \cite{Ramakrishnan2021a} and Gibson\cite{Xia2018} datasets. Both Gibson and HM3D contain high-resolution photorealistic 3D reconstructions of real-world environments.
For Gibson, we use 25 train and 5 val scenes from the Gibson tiny split which already has associated semantic annotations. For HM3D, we use the standard splits of 75 train and 20 val scenes with habitat format.
There are 6 object goal categories defined \cite{Chaplot2020b}: chair, couch, potted plant, bed, toilet and tv.
To provide more semantic relevance, we also detect other most frequent 7 categories which are common between Gibson and HM3D datasets during the navigation.

\subsubsection{Experiment Details}

We evaluate our methods in the 3D indoor simulator Habitat platform \cite{Savva2019}.
The observation space contains $480\times640$ RGBD images, base odometry sensor, and goal object category represented as an integer. For the semantic segmentation, we use the pretrained RedNet model \cite{Jiang2018a} finetuned as \cite{Ye2021a} to predict all the categories.
We build our model based on the publicly available code of \cite{Chaplot2020b} and \cite{Huang2022}, and use the PyTorch framework. For training the policy using PPO algorithm, we adapte Adam optimizer with a learning rate of 0.000025, a discount factor of $\gamma$ = 0.99. The global map size is $96\times96$ $m$, the local map size is $48\times48$ $m$, and the map resolution is 0.05 $m$. The weight $\lambda_{C U} $ is 0.5. We train our model for over 2 million frames on NVIDIA V100 32GB GPU.

\subsubsection{Evaluation Metrics}

We follow \cite{Anderson2018} to evaluate our method using Success Rate, Success weighted by Path Length (SPL), and Distance to Goal (DTG). SR is defined as $\frac{1}{N} \sum_{i=1}^{N} S_{i}$, and SPL is defined as $\frac{1}{N} \sum_{i=1}^{N} S_{i} \frac{l_{i}}{\max \left(l_{i}, p_{i}\right)}$, where $N$ is the number of episodes, $S_{i}$ is 1 if the episode is successful, else is 0, $l_{i}$ is the shortest trajectory length between the start position and one of the success position, $p_{i}$ is the trajectory length of the current episode $i$. The DTG is the distance between the agent and the target goal when the episode ends.

\subsubsection{Baselines}

To assess the navigation performance of our model on the evaluation datasets, a few baselines are considered:

\begin{itemize}

	\item \textbf{Randomly Walking: }The agent randomly samples an action from the action space $\mathcal{A}$ with a uniform distribution at each step.
	\item \textbf{Frontier-based Policy \cite{Yamauchi1997}: }This baseline uses classical robotics pipeline for mapping followed by classical frontier-based exploration algorithm. Whenever the goal object category is detected, the local policy tries to go towards the object using an analytical planner.
	\item \textbf{Randomly Sampling on Map: }We build the explored map firstly from the observation of the agent. Secondly, the long-term goal is sampled from the map with a uniform distribution at each step. After getting the long term goal, the final action can be selected by the local planner method.
	\item \textbf{SemExp \cite{Chaplot2020b}: }We follow \cite{Chaplot2020b} as the baseline to explore and search for the target using semantic priors.

\end{itemize}


\begin{figure*}[htbp]
	\centering
	\subfloat[step 1]
	{
		\includegraphics[width=4.2cm]{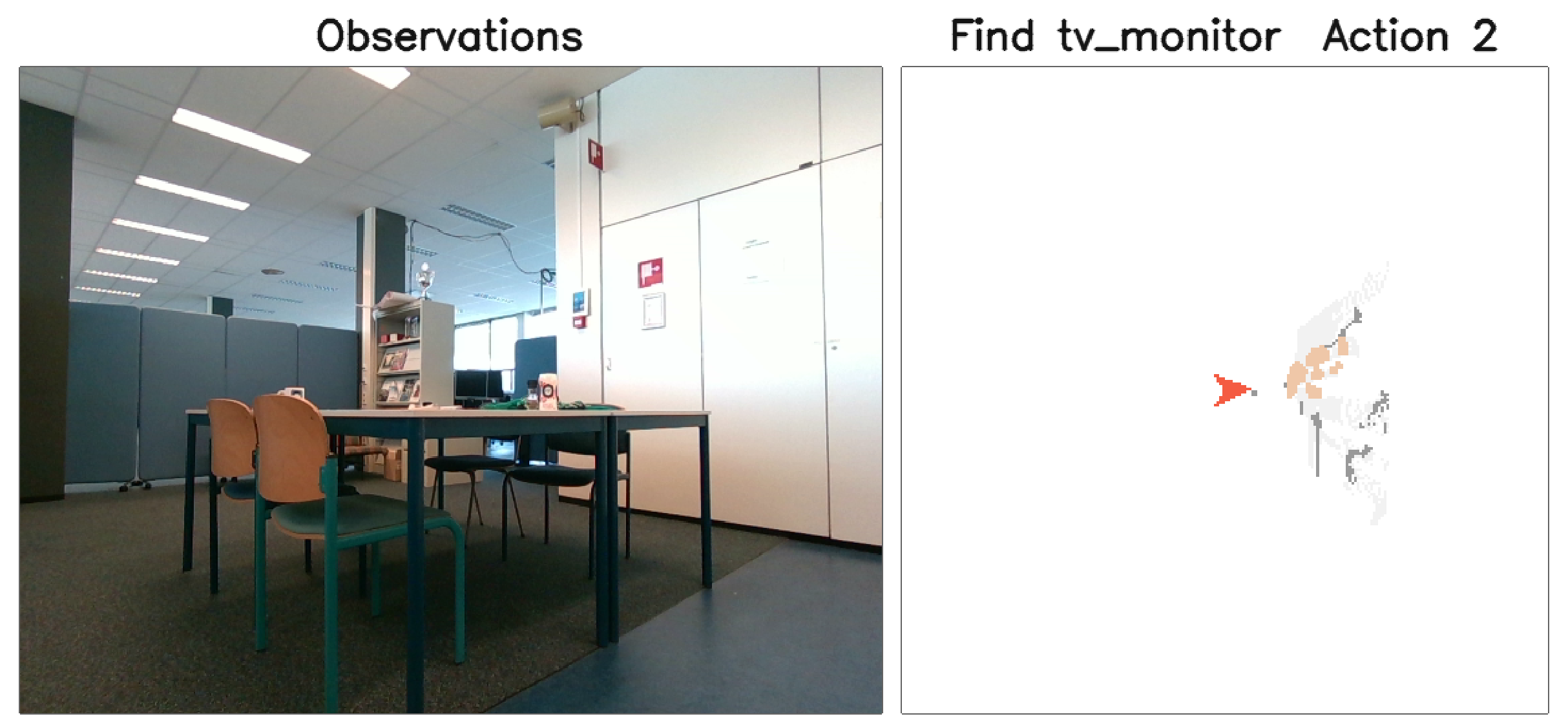}
	}
	\subfloat[step 30]
	{
		\includegraphics[width=4.2cm]{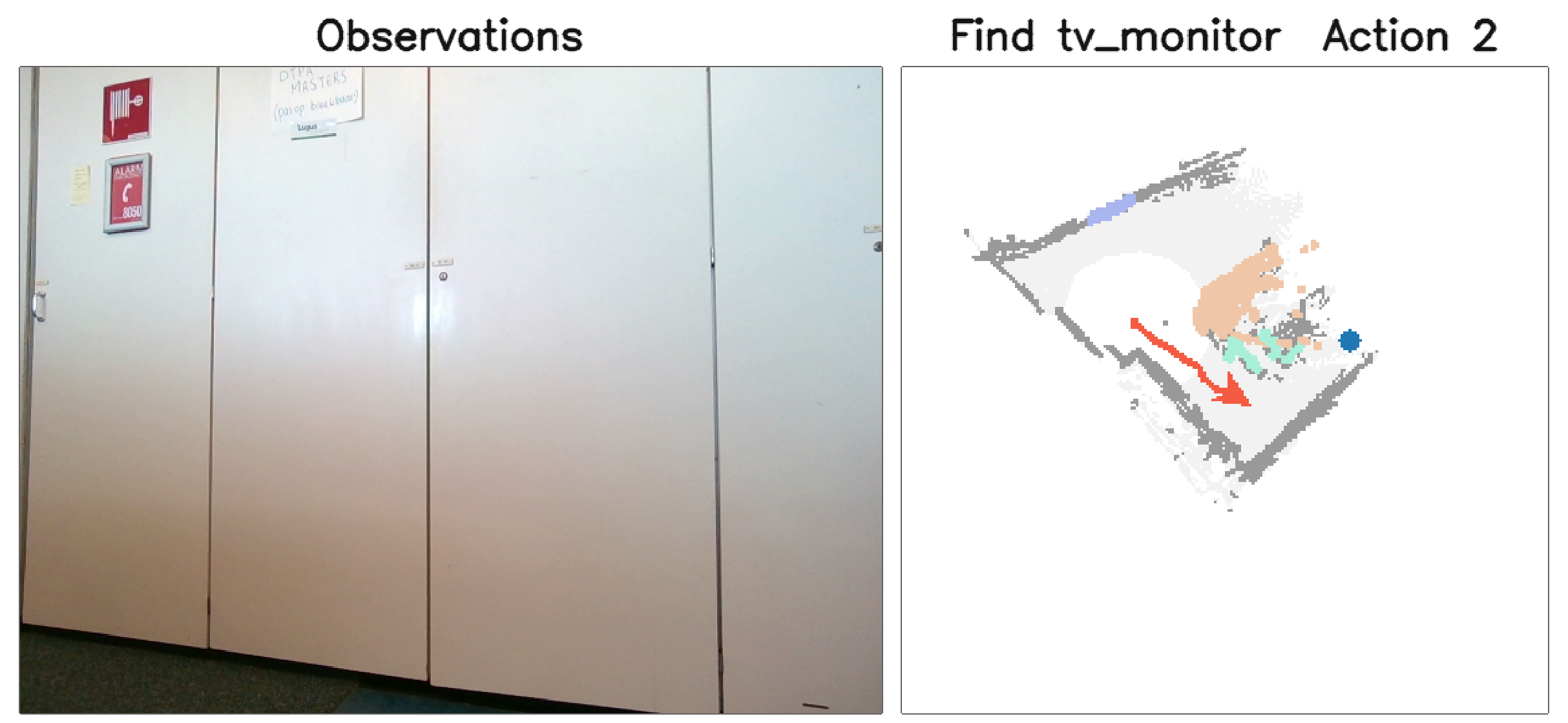}
	}
	\subfloat[step 50]
	{
		\includegraphics[width=4.2cm]{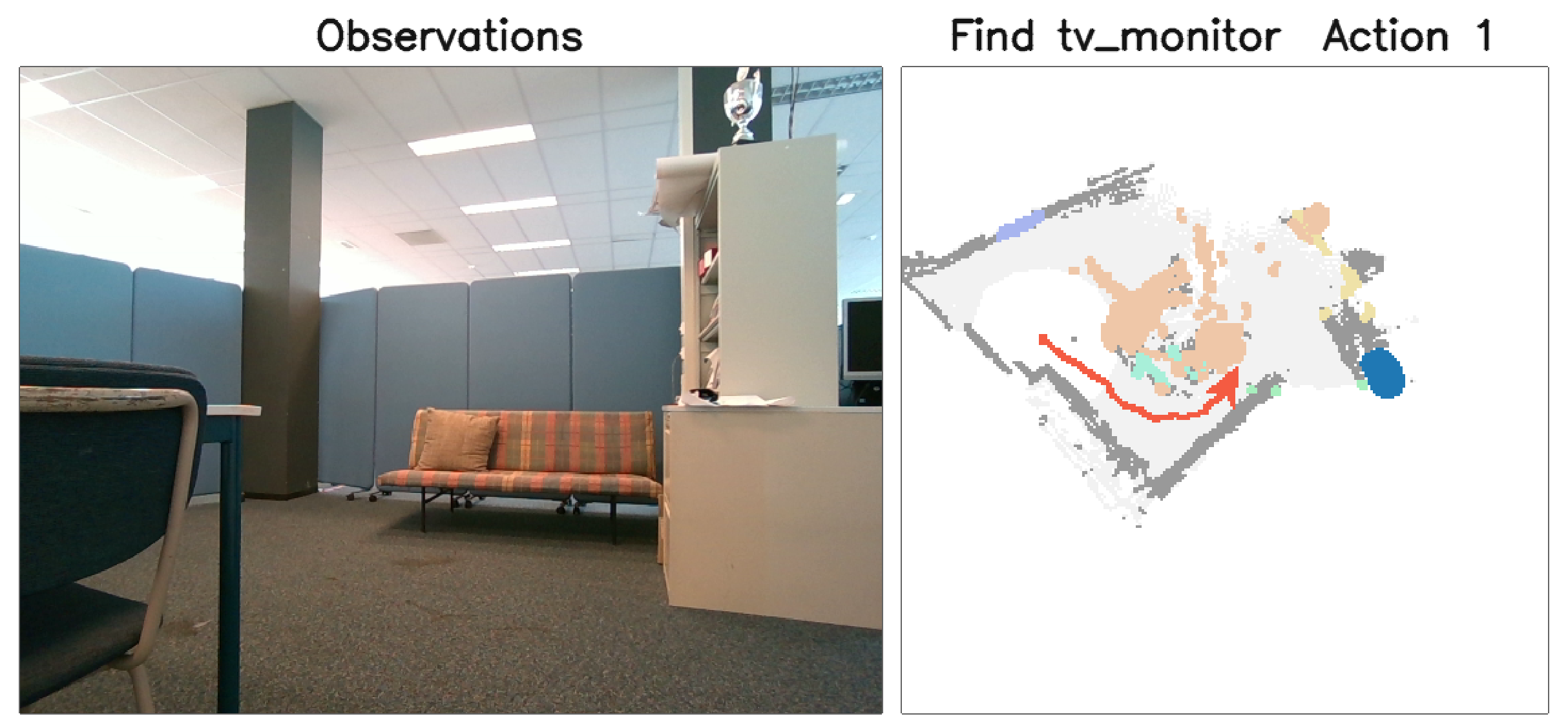}
	}
	\subfloat[step 72]
	{
		\includegraphics[width=4.2cm]{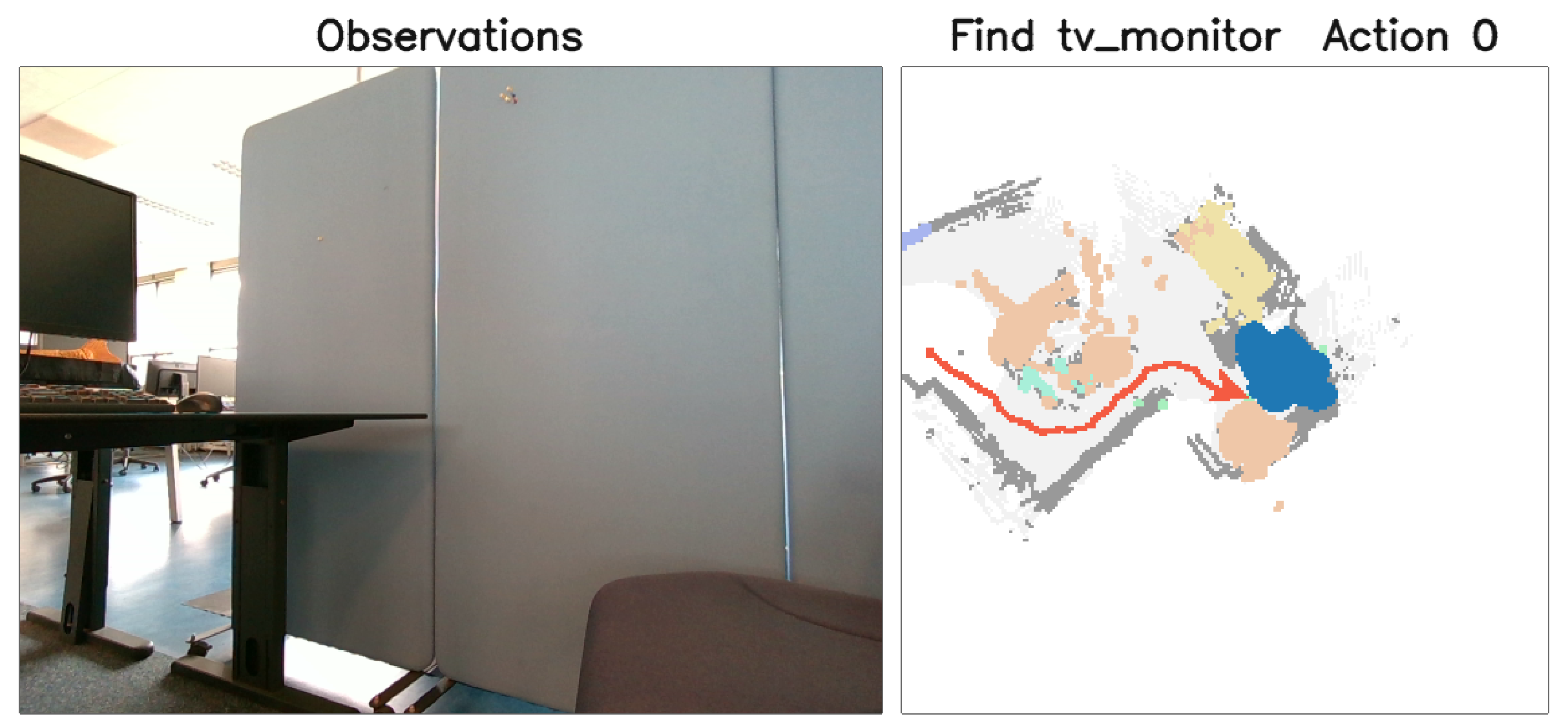}
	}
	\caption{The process of visual target navigation experiment in the real world to find a monitor.}
	\label{fig:real_episode}
\end{figure*}

\subsubsection{Result and Discussion}

The quantitative results of the comparison study are reported in TABLE \ref{tab:performance_comparison}.
From the table we can find that random walking almost fails in all the episodes, which illustrates that each episode of the datasets is not easy to get success using randomly discrete actions.
However, when we randomly sample the long-term goal using the map-based framework, the performance is even better than the classical frontier-based method \cite{Yamauchi1997}. This suggests the great advantage of the map-based method is the long-term goal can help the robot explore the environment roughly and quickly.
The great improvement in SemExp \cite{Chaplot2020b} shows the importance of the semantic information in the scene is that the long-term goal can be selected more efficiently based on the semantic relevance.
As can be seen that our method outperforms all the baselines by a considerable margin consistently across both the datasets, which achieves 71.5 \% and 53.8 \% success rate on Gibson/HM3D, and improves a lot over the SemExp baseline \cite{Chaplot2020b}, which shows the efficiency of our framework. The process of finding a toilet is shown in Fig \ref{fig:sim_episode}.

\subsubsection{Ablation study}

To understand the importance of the different modules in our framework, we consider the following ablations based on the HM3D dataset:

\begin{itemize}

	\item \textbf{FSE w.o. Frontier Map: } The importance of the semantic map has been proved in \cite{Chaplot2020b}. So we replace the semantic and frontier maps with the semantic-only map. The policy is still retrained with the same reward function and samples a long-term goal from frontiers as before.

	\item \textbf{FSE w.o. Goal Policy: } Our policy is trained to select the best frontiers as the long-term goal. To verify the performance, we replace our learned frontier semantic policy with a goal-agnostic policy \cite{Chaplot2020} which samples the point on the frontier cell closest to the agent.

	\item \textbf{FSE w. GT SemSeg: } We use the ground-truth semantic sensor in the Habitat simulator to replace the semantic segmentation in this framework.

\end{itemize}

\begin{table}[htbp]
	\centering
	\fontsize{10}{9}\selectfont
	\begin{threeparttable}
		\caption{Results of Ablation Study in HM3D.}
		\label{tab:ablation_study}
		\begin{tabular}{cccc}
			\toprule
			Method                & Success & SPL   & DTG(m)       \cr
			\midrule
			FSE w.o. Frontier Map & 0.470   & 0.218 & 4.725         \cr
			FSE w.o. FS Policy    & 0.465   & 0.217 & 3.893           \cr
			FSE                   & 0.538   & 0.246 & 3.745         \cr
			FSE w. GT SemSeg      & 0.640   & 0.329 & 3.667    \cr
			\bottomrule
		\end{tabular}
	\end{threeparttable}
\end{table}

The importance of the frontier map and frontier semantic policy in the visual navigation task are remarkable in TABLE \ref{tab:ablation_study}. Especially, we can notice that the DTS of the FSE w.o. Frontier Map is larger than the FES w.o. FS Policy compared with Success and SPL. It indicates that the frontier map is an important feature for the policy to select the long-term goal, and it is possible that the policy randomly samples from frontier cells if there is no correlation between input feature and action space. In FSE w.o. FS Policy, a certain phase consistency should exist in the goal-agnostic policy as the DTS decreased.

The performance of the proposed model is still far from perfect. The error is also needed to analyze for future improvement. We can see that the FSE w. GT SemSeg improves about 10\% in the Success and SPL to our method, which means that the accuracy of the semantic segmentation make a great influence on the results. From our observation, another important factor is the multi-floor scene in the HM3D dataset that the start position of the robot and the goal are not on the same floor. This is a great challenge for the 2D map-based method, because going upstairs or downstairs is difficult for the robot and the multi-floor map would destroy the continuity of the map. 3D map-based exploration may be a great way to solve this problem.

\subsection{Real World Experiment}

We use ROS\cite{Quigley2009} and the Jackal Robot hardware platform with Realsense D455 camera and Ouster lidar to deploy the frontier semantic policy in the real world.

To bridge the gap between simulation and real world, we keep all sensor data as similar as possible to the simulation environment.
Firstly, we set the same height of the RGB-D camera and limit the same range of the depth image for the robot as the setting in the simulation.
To get a more accurate location of the robot, the lidar is also used to build the geometry map in real-time to correct the location.
Then the RGB-D images, location, and object category would be sent into our model and the action would be output. From Fig \ref{fig:real_episode} we can see that the long-term goal (blue spot) selected by our model can guide the robot to explore the environment and search for the goal efficiently.

The noise of the images and location is the main gap between simulation and the real world. The depth image is influenced by the illumination and the limitation of the hardware. Even though we equip with many depth image filters, there are still some wrong projections about the chair in Fig \ref{fig:real_episode} (orange area). On the other hand, we can't get enough accurate location using lidar and odometry compared with simulation, which causes the noisy points around the wall or other objects. But the advantage of the map-based framework in the real-world transfer is that the input of our model is the scene map rather than the direct images with noise, which can significantly decrease the influence of the noise on our model. So we can deploy our model on the robot platform without much fine-tuning.

\section{CONCLUSIONS}

In this paper, we present a frontier semantic exploration framework to tackle the visual target navigation task in an unknown environment. The frontier and semantic maps are used to select the long-term goal from the frontier cells. By implementing the experiments in Gibson and HM3D scenes, we show that our method significantly improves the performance of success rate and efficiency.
Ablation studies show that the proposed model learns frontier priors which leads to more efficient exploration. Real-world experiment shows the advantage of the map-based framework and the applicability of our method.

\bibliographystyle{ieeetr}
\bibliography{bib/library.bib}

\begin{thebibliography}{10}

\bibitem{Zhu2017}
Y.~Zhu, R.~Mottaghi, E.~Kolve, J.~J. Lim, A.~Gupta, L.~Fei-Fei, and A.~Farhadi,
  ``{Target-driven visual navigation in indoor scenes using deep reinforcement
  learning},'' in {\em 2017 IEEE International Conference on Robotics and
  Automation (ICRA)}, pp.~3357--3364, IEEE, may 2017.

\bibitem{Yang2019}
W.~Yang, X.~Wang, A.~Farhadi, A.~Gupta, and R.~Mottaghi, ``{Visual semantic
  navigation using scene priors},'' in {\em 7th International Conference on
  Learning Representations, ICLR 2019}, pp.~1--14, 2019.

\bibitem{Lyu2022}
Y.~Lyu, Y.~Shi, and X.~Zhang, ``{Improving Target-driven Visual Navigation with
  Attention on 3D Spatial Relationships},'' {\em Neural Processing Letters},
  2022.

\bibitem{Druon2020}
R.~Druon, Y.~Yoshiyasu, A.~Kanezaki, and A.~Watt, ``{Visual object search by
  learning spatial context},'' {\em IEEE Robotics and Automation Letters},
  vol.~5, no.~2, pp.~1279--1286, 2020.

\bibitem{Ye2021a}
J.~Ye, D.~Batra, A.~Das, and E.~Wijmans, ``{Auxiliary Tasks and Exploration
  Enable ObjectGoal Navigation},'' {\em Proceedings of the IEEE International
  Conference on Computer Vision}, pp.~16097--16106, 2021.

\bibitem{Chaplot2020b}
D.~S. Chaplot, D.~Gandhi, A.~Gupta, and R.~Salakhutdinov, ``{Object goal
  navigation using goal-oriented semantic exploration},'' {\em Advances in
  Neural Information Processing Systems}, vol.~2020-Decem, no.~NeurIPS,
  pp.~1--12, 2020.

\bibitem{Savva2019}
M.~Savva, A.~Kadian, O.~Maksymets, Y.~Zhao, E.~Wijmans, B.~Jain, J.~Straub,
  J.~Liu, V.~Koltun, J.~Malik, D.~Parikh, and D.~Batra, ``{Habitat: A Platform
  for Embodied AI Research},'' in {\em 2019 IEEE/CVF International Conference
  on Computer Vision (ICCV)}, vol.~2019-Octob, pp.~9338--9346, IEEE, oct 2019.

\bibitem{Xia2018}
F.~Xia, A.~R. Zamir, Z.~He, A.~Sax, J.~Malik, and S.~Savarese, ``{Gibson Env:
  Real-World Perception for Embodied Agents},'' in {\em 2018 IEEE/CVF
  Conference on Computer Vision and Pattern Recognition}, pp.~9068--9079, IEEE,
  jun 2018.

\bibitem{Huang2022}
S.~Huang and S.~Onta{\~{n}}{\'{o}}n, ``{A Closer Look at Invalid Action Masking
  in Policy Gradient Algorithms},'' {\em The International FLAIRS Conference
  Proceedings}, vol.~35, may 2022.

\bibitem{Ramakrishnan2021a}
S.~K. Ramakrishnan, A.~Gokaslan, E.~Wijmans, O.~Maksymets, A.~Clegg, J.~Turner,
  E.~Undersander, W.~Galuba, A.~Westbury, A.~X. Chang, M.~Savva, Y.~Zhao, and
  D.~Batra, ``{Habitat-Matterport 3D Dataset (HM3D): 1000 Large-scale 3D
  Environments for Embodied AI},'' {\em arXiv}, sep 2021.

\bibitem{Labbe2013}
M.~Labbe and F.~Michaud, ``{Appearance-Based Loop Closure Detection for Online
  Large-Scale and Long-Term Operation},'' {\em IEEE Transactions on Robotics},
  vol.~29, pp.~734--745, jun 2013.

\bibitem{Nakajima2019}
Y.~Nakajima and H.~Saito, ``{Efficient object-oriented semantic mapping with
  object detector},'' {\em IEEE Access}, vol.~7, pp.~3206--3213, 2019.

\bibitem{Sun2018}
H.~Sun, Z.~Meng, P.~Y. Tao, and M.~H. Ang, ``{Scene Recognition and Object
  Detection in a Unified Convolutional Neural Network on a Mobile
  Manipulator},'' in {\em 2018 IEEE International Conference on Robotics and
  Automation (ICRA)}, pp.~1--5, IEEE, may 2018.

\bibitem{Grinvald2019}
M.~Grinvald, F.~Furrer, T.~Novkovic, J.~J. Chung, C.~Cadena, R.~Siegwart, and
  J.~Nieto, ``{Volumetric Instance-Aware Semantic Mapping and 3D Object
  Discovery},'' {\em IEEE Robotics and Automation Letters}, vol.~4,
  pp.~3037--3044, jul 2019.

\bibitem{He2016}
K.~He, X.~Zhang, S.~Ren, and J.~Sun, ``{Deep Residual Learning for Image
  Recognition},'' in {\em 2016 IEEE Conference on Computer Vision and Pattern
  Recognition (CVPR)}, vol.~2016-Decem, pp.~770--778, IEEE, jun 2016.

\bibitem{Mnih2013}
V.~Mnih, A.~{Puigdom{\`{e}}nech Badia}, M.~Mirza, T.~Harley, T.~{P. Lillicrap},
  D.~Silver, and K.~Kavukcuoglu, ``{Asynchronous Methods for Deep Reinforcement
  Learning Volodymyr},'' {\em International Conference on Machine Learning},
  vol.~48, 2013.

\bibitem{Kipf2017}
T.~N. Kipf and M.~Welling, ``{Semi-Supervised Classification with Graph
  Convolutional Networks},'' {\em arXiv}, pp.~1--14, sep 2016.

\bibitem{Du2020}
H.~Du, X.~Yu, and L.~Zheng, ``{Learning Object Relation Graph and Tentative
  Policy for Visual Navigation},'' {\em Lecture Notes in Computer Science
  (including subseries Lecture Notes in Artificial Intelligence and Lecture
  Notes in Bioinformatics)}, vol.~12352 LNCS, pp.~19--34, 2020.

\bibitem{Maksymets2021}
O.~Maksymets, V.~Cartillier, A.~Gokaslan, E.~Wijmans, W.~Galuba, S.~Lee, and
  D.~Batra, ``{THDA: Treasure Hunt Data Augmentation for Semantic
  Navigation},'' {\em Proceedings of the IEEE International Conference on
  Computer Vision}, pp.~15354--15363, 2021.

\bibitem{Wijmans2019}
E.~Wijmans, A.~Kadian, A.~Morcos, S.~Lee, I.~Essa, D.~Parikh, M.~Savva, and
  D.~Batra, ``{DD-PPO: Learning Near-Perfect PointGoal Navigators from 2.5
  Billion Frames},'' {\em arXiv}, nov 2019.

\bibitem{Mousavian}
A.~Mousavian, A.~Toshev, M.~Fiser, J.~Kosecka, A.~Wahid, and J.~Davidson,
  ``{Visual Representations for Semantic Target Driven Navigation},'' in {\em
  2019 International Conference on Robotics and Automation (ICRA)},
  pp.~8846--8852, IEEE, may 2019.

\bibitem{Fang2019}
K.~Fang, A.~Toshev, L.~Fei-Fei, and S.~Savarese, ``{Scene Memory Transformer
  for Embodied Agents in Long-Horizon Tasks},'' in {\em 2019 IEEE/CVF
  Conference on Computer Vision and Pattern Recognition (CVPR)},
  vol.~2019-June, pp.~538--547, IEEE, jun 2019.

\bibitem{Chaplot2020}
D.~S. Chaplot, D.~Gandhi, S.~Gupta, A.~Gupta, and R.~Salakhutdinov, ``{Learning
  to Explore using Active Neural SLAM},'' in {\em International Conference on
  Learning Representations (ICLR)}, apr 2020.

\bibitem{Chaplot2020a}
D.~S. Chaplot, R.~Salakhutdinov, A.~Gupta, and S.~Gupta, ``{Neural Topological
  SLAM for Visual Navigation},'' in {\em Proceedings of the IEEE/CVF Conference
  on Computer Vision and Pattern Recognition (CVPR)}, pp.~12875--12884, 2020.

\bibitem{Chang2020}
M.~Chang, A.~Gupta, and S.~Gupta, ``{Semantic visual navigation by watching
  YouTube videos},'' {\em Advances in Neural Information Processing Systems},
  vol.~2020-Decem, no.~NeurIPS, 2020.

\bibitem{Ramakrishnan2022}
S.~K. Ramakrishnan, D.~S. Chaplot, Z.~Al-Halah, J.~Malik, and K.~Grauman,
  ``{PONI: Potential Functions for ObjectGoal Navigation with Interaction-free
  Learning},'' 2022.

\bibitem{Niroui2019}
F.~Niroui, K.~Zhang, Z.~Kashino, and G.~Nejat, ``{Deep Reinforcement Learning
  Robot for Search and Rescue Applications: Exploration in Unknown Cluttered
  Environments},'' {\em IEEE Robotics and Automation Letters}, vol.~4, no.~2,
  pp.~610--617, 2019.

\bibitem{Yamauchi1997}
B.~Yamauchi, ``{Frontier-based approach for autonomous exploration},'' {\em
  Proceedings of IEEE International Symposium on Computational Intelligence in
  Robotics and Automation, CIRA}, pp.~146--151, 1997.

\bibitem{Nanni2011}
L.~P. Kaelbling, M.~L. Littman, and A.~R. Cassandra, ``{Planning and acting in
  partially observable stochastic domains},'' {\em Artificial Intelligence},
  vol.~101, pp.~99--134, may 1998.

\bibitem{Julia2012}
M.~Juli{\'{a}}, A.~Gil, and O.~Reinoso, ``{A comparison of path planning
  strategies for autonomous exploration and mapping of unknown environments},''
  {\em Autonomous Robots}, vol.~33, no.~4, pp.~427--444, 2012.

\bibitem{Schulman2017}
J.~Schulman, F.~Wolski, P.~Dhariwal, A.~Radford, and O.~Klimov, ``{Proximal
  Policy Optimization Algorithms},'' {\em ArXiv}, 2017.

\bibitem{Sethian1996}
J.~A. Sethian, ``{A fast marching level set method for monotonically advancing
  fronts},'' {\em Proceedings of the National Academy of Sciences of the United
  States of America}, vol.~93, no.~4, pp.~1591--1595, 1996.

\bibitem{Anderson2018}
P.~Anderson, A.~Chang, D.~S. Chaplot, A.~Dosovitskiy, S.~Gupta, V.~Koltun,
  J.~Kosecka, J.~Malik, R.~Mottaghi, M.~Savva, and A.~R. Zamir, ``{On
  Evaluation of Embodied Navigation Agents},'' {\em arXiv}, jul 2018.

\bibitem{Ramakrishnan2021}
S.~K. Ramakrishnan, D.~Jayaraman, and K.~Grauman, ``{An Exploration of Embodied
  Visual Exploration},'' {\em International Journal of Computer Vision},
  vol.~129, pp.~1616--1649, may 2021.

\bibitem{Jiang2018a}
J.~Jiang, L.~Zheng, F.~Luo, and Z.~Zhang, ``{RedNet: Residual Encoder-Decoder
  Network for indoor RGB-D Semantic Segmentation},'' {\em arXiv}, jun 2018.

\bibitem{Quigley2009}
M.~Quigley, K.~Conley, B.~Gerkey, J.~Faust, T.~Foote, J.~Leibs, R.~Wheeler, and
  A.~Y. Ng, ``{ROS: an open-source Robot Operating System},'' in {\em ICRA
  workshop on open source software}, vol.~3, 2009.

\end{thebibliography}

\end{document}